\documentclass[11pt]{article}
\usepackage[preprint]{acl}
\usepackage{times,latexsym}
\usepackage[T1]{fontenc}
\usepackage[utf8]{inputenc}
\usepackage{microtype,inconsolata}
\usepackage{graphicx,amsmath,amssymb,booktabs,longtable,array}
\usepackage{colortbl}
\usepackage{float}
\usepackage{placeins}
\usepackage{dblfloatfix}
\graphicspath{{figures/}}
\title{WMIP: A Working-Memory-Inspired Programmable Probe of Cumulative State Tracking in Large Language Models}
\author{%
  Dengzhe Hou\textsuperscript{1,*}\hspace{1.5em}
  Lingyu Jiang\textsuperscript{1}\hspace{1.5em}
  Deng Li\textsuperscript{2}\hspace{1.5em}
  Zirui Li\textsuperscript{1}\\[2pt]
  \bf Fangzhou Lin\textsuperscript{1,3,4}\hspace{1.5em}
  Kazunori D Yamada\textsuperscript{1}\\[6pt]
  \normalfont\textsuperscript{1}Tohoku University\hspace{1em}
  \textsuperscript{2}Lappeenranta-Lahti University of Technology LUT\\[1pt]
  \textsuperscript{3}Texas A\&M University\hspace{1em}
  \textsuperscript{4}Worcester Polytechnic Institute\\[2pt]
  {\small \textsuperscript{*}Corresponding author.\quad Project:
  \href{https://github.com/dengzhe-hou/WMIP}{\nolinkurl{github.com/dengzhe-hou/WMIP}}}
}
\begin{document}
\raggedbottom
\setlength{\textfloatsep}{12pt plus 2pt minus 2pt}
\setlength{\dbltextfloatsep}{12pt plus 2pt minus 2pt}
\maketitle
\begin{abstract}
Human working memory lets people maintain and update information.
Large language models (LLMs) face a related demand when tracking updates within a single prompt without writing out intermediate states, yet existing tests differ in their constraints on written steps.
We introduce a working-memory-inspired probe of cumulative state tracking with program-computed answers, 3 to 100 updates, and final-state, intermediate-state, or multi-target queries.
It compares protocols on identical items.
Across 18 LLMs, accuracy is 84.9\% on single-step arithmetic but 31.7\% on core items requiring one number after three to seven updates.
From 50 updates on, no model exceeds 10\% accuracy.
Core accuracy reaches 88.0\% under a larger budget and instructions to write steps and place the answer on the last line, mainly due to the instruction changes rather than the budget.
Core accuracy also correlates positively with public Berkeley Function Calling Leaderboard scores.
These findings show that strong single-step arithmetic does not ensure reliable cumulative tracking and that output protocols substantially change measured accuracy.
We release the generator, items, scoring code, and responses for reproducible comparisons across depths, query types, and protocols.
\end{abstract}
\section{Introduction}
\label{sec:intro}

Keeping a running score in one's head relies on working memory, which holds information briefly and updates it as changes arrive~\citep{baddeley1974working,miyake2000unity}.
Large language models (LLMs) meet a similar demand when a prompt lists an initial state and a series of updates and asks for the result without written steps~\citep[cf.][]{nye2021show,wei2022chain}.
Such state tracking underlies many uses of LLMs, from understanding narratives and instructions to stateful, multi-turn tool use~\citep{entitytracking2023,patil2025bfcl}.
Much as people write down information they need to remember~\citep{risko2016cognitive}, a model's written steps record intermediate states, which removes this demand~\citep{nye2021show}.
Testing this demand therefore requires controlling how prompt wording~\citep{mizrahi2024state}, output budgets, and scoring rules constrain written steps.

Existing state-tracking tests share this structure but constrain written steps differently.
Minerva, for example, allows 4,096 output tokens, so models can write out the operations~\citep{xia2025minerva}.
These tests may therefore not all pose this demand.

\begin{figure*}[t]
\centering
\includegraphics[width=\textwidth]{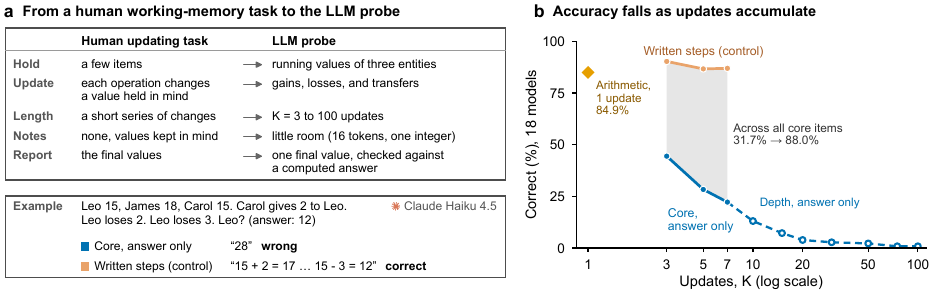}
\caption{\textbf{In a working-memory-inspired probe, LLM accuracy falls as updates accumulate when models answer without written steps.} (a) The probe mirrors human working-memory updating tasks~\citep{miyake2000unity}, in which an update can transform a held value rather than only replace it~\citep{ecker2010components}. The core protocol restricts written steps, paralleling participants' need to keep values in mind, but keeps the full update sequence visible. In the abridged example (full prompt in Section~\ref{sec:generation}), Claude Haiku 4.5 answers incorrectly under the core protocol and correctly under the written-steps protocol. (b) Accuracy pooled over the 18 models. Solid lines compare protocols on the 60 core items ($K=3$, 5, 7). The dashed line extends the core protocol to depth items ($K\ge10$) from the same generator. The diamond marks single-step arithmetic on separate items.}
\label{fig:framework}
\end{figure*}

We present a working-memory-inspired probe of cumulative state tracking (Figure~\ref{fig:framework}).
Its generator sets the number of updates, and its protocols set how much room is left for written steps.
Under the core protocol, a 16-token budget and the one-integer rule leave little room for written steps.

Without written steps, all 18 models answer the core items less accurately than single-step arithmetic, and accuracy falls as sequences grow.
Asking for written steps~\citep{kojima2022large} raises core accuracy from 31.7\% to 88.0\%, so protocol choices change what is measured.
Core accuracy also correlates positively with public Berkeley Function Calling Leaderboard (BFCL) scores~\citep{patil2025bfcl} (Section~\ref{sec:association}).

We make three contributions.
\begin{itemize}
\item A generator with adjustable depth, executable answers, and final-state, intermediate-state, and multi-target queries, released with items, code, and responses.
\item Same-item comparisons showing substantial protocol effects on measured accuracy, with sensitivity analyses across wordings, seeds, and depths.
\item Positive BFCL associations across 18 models, compared with those of single-step arithmetic and cancellation controls.
\end{itemize}

\section{Related Work}
\label{sec:related}

Entity tracking moves objects among boxes~\citep{entitytracking2023}, and BIG-Bench Hard tracks pairwise object swaps with multiple-choice answers~\citep{suzgun2023challenging}. Our core instead applies integer updates to three named entities and requests one free-response integer. Minerva's quantity-state task is closest in content but allows 4,096 output tokens, enough to work through updates in writing~\citep{xia2025minerva}. GSM-Infinite varies operation counts but requests step-by-step reasoning~\citep{zhou2025gsminfinite}. Our core allows 16 tokens and scores exactly one integer. Written-steps and large-budget protocols relax these limits on the same items.

LLM working-memory studies use match detection on $n$-back tasks~\citep{gong2024working,zhang2024working}, and tested models meet or exceed human norms on span tasks~\citep{de2025strong}. \citet{wang2025proactive} ask for the last value of a repeatedly overwritten key. Unlike human operation span~\citep{conway2005working}, our probe keeps every initial value and update visible. Working memory therefore motivates the demand tested here, without establishing equivalence to human span tasks. Representation probes and interventions can distinguish state-tracking algorithms~\citep{li2025howtrack,tang2026entity}, but final answers cannot show whether models maintain or recompute states.

\begin{figure*}[t]
\centering
\includegraphics[width=\textwidth]{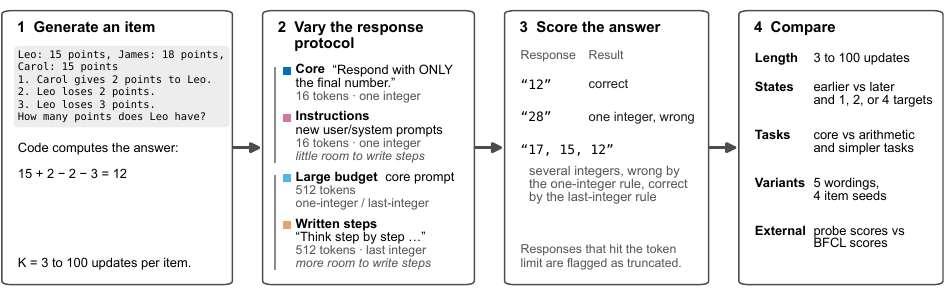}
\caption{Probe workflow. (1) Code generates items and computes reference answers. (2) Four response protocols are applied to the same core items. (3) Responses are scored and truncation is flagged. (4) Performance is compared across conditions and related to BFCL scores. The response 28 is Claude Haiku 4.5's recorded response (Figure~\ref{fig:framework}a). The other two illustrate the scoring rules.}
\label{fig:overview}
\end{figure*}

Benchmark scores depend on paraphrased instructions and task-preserving formatting~\citep{mizrahi2024state,sclar2024quantifying}. Output-format restrictions can affect reasoning accuracy~\citep{tam2024speak}. Motivated by measurement-validity recommendations~\citep{bean2025measuring}, we report integer compliance, truncation, and seed and wording sensitivity alongside accuracy. BFCL evaluates function calling, including stateful, multi-step interactions~\citep{patil2025bfcl}, and shares no items with the probe. Comparing the core's BFCL association with those of simpler controls follows discriminant-validation logic~\citep{campbell1959convergent}. Appendix~\ref{app:related} provides additional task and methodological comparisons.

\section{Probe and Evaluation Design}
\label{sec:method}

\subsection{Generating state-tracking tasks}
\label{sec:generation}
Core items initialize three named entities with integers, list $K$ ordered gains, losses, or transfers between them, and query one entity's final value (Figure~\ref{fig:overview}).
Initial values range from 5 to 20, gains from 1 to 10, losses from 1 to 5, and transfers from 1 to 3.
The reference answer is the queried entity's initial value plus its signed changes, so update order does not change it. A short program computes it and counts the updates to that entity (median 2 at $K=3$ and 45.5 at $K=100$).
A fixed redraw rule excludes core and depth items where that entity is never updated or finishes at its initial value, preventing correct answers from copying the initial value.

Each item is one user message.
Although the prompt says the initial state cannot be consulted again, the full sequence remains in context.
The core prompt ends with \emph{Respond with ONLY the final number.} and allows 16 output tokens.
One complete core prompt reads as follows.
{\setlength{\leftmargini}{0.6em}\begin{quote}\small
You will track a sequence of point updates. You cannot refer back to the initial state after reading it once.

Initial state:\newline Leo: 15 points, James: 18 points, Carol: 15 points

Operations (apply in order):\newline   1. Carol gives 2 points to Leo.\newline   2. Leo loses 2 points.\newline   3. Leo loses 3 points.

After all operations, how many points does Leo have?

Respond with ONLY the final number.
\end{quote}
The reference answer is 12.
}
Appendix~\ref{app:prompts} details task generation and prompt variants.

\subsection{Core bank and comparison tasks}
\label{sec:bank}
\begin{table}[!b]
\centering\small
\begin{tabular}{@{}>{\raggedright\arraybackslash}p{1.7cm}@{\hspace{5pt}}>{\raggedright\arraybackslash}p{3.25cm}@{\hspace{2pt}}r@{\hspace{8pt}}r@{}}
\toprule
Block & Purpose & Requests & Tokens \\
\midrule
\multicolumn{4}{@{}l}{\emph{Main bank}} \\
Core & Primary score & 60 & 16 \\
Depth & Effect of sequence length & 140 & 16 \\
Arithmetic & Arithmetic alone & 90 & 16 \\
Cancellation & No net change & 100 & 16 \\
Assignment & Non-arithmetic tracking & 90 & 16 \\
Sets/counts & Set tracking and counting & 90 & 48 or 16 \\
Wordings & Wording sensitivity & 150 & 16 \\
Instructions & Instruction sensitivity & 60 & 16 \\
Written steps & Steps + final-line answer & 60 & 512 \\
\midrule
\multicolumn{4}{@{}l}{\emph{Further blocks}} \\
Intermediate states & Test end-state shortcuts & 60 & 16 \\
Multi-target & Tracking and readout & 120 & 48 \\
Large budget & Effect of output budget & 60 & 512 \\
\bottomrule
\end{tabular}
\caption{Requests per model and output-token limits by block (sets 48, counts 16).}
\label{tab:bank-summary}
\end{table}
Each model receives 1,080 requests across nine main-bank blocks (840 requests) and three further blocks (Table~\ref{tab:bank-summary}), with repeated items across wordings and protocols.
Core accuracy uses 60 items, five at each depth ($K=3,5,7$) for each of four generation seeds.
The same generator produces 20 items per depth at $K=10,15,20,30,50,75,100$.
Depths use different instances, not longer prefixes of one sequence.

Single-step arithmetic asks for one update across three domains and magnitude ranges.
Cancellation items list adjacent updates with no net change at $K=2,4,6,8,12$.
Assignment items directly assign colors, locations, or statuses to three entities.
Set and count items comprise 60 permission and inventory set-update items and 30 meeting-count items with increments or decrements of one.
Both blocks have ten items per domain at each of $K=3,5,7$.
Assignment differs from the core in answer type. Sets and counts also differ in entity count and some output limits.

\begin{figure*}[!b]
\centering
\includegraphics[width=\textwidth]{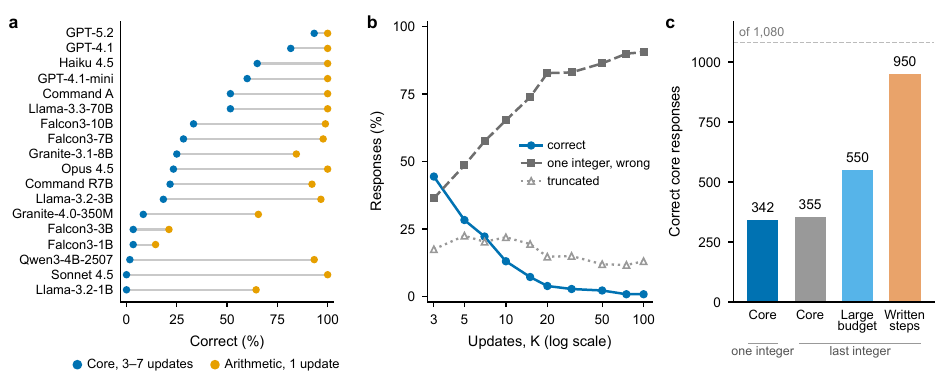}
\caption{Main results for 18 models. (a) Accuracy on 60 core and 90 single-step arithmetic items, with core scores from 0/60 to 56/60. (b) Response rates by depth $K$. At each depth, all 18 models answer the same 20 items (360 responses). Items are regenerated across depths. Surrounding text is allowed under the one-integer rule. Truncation can overlap the response categories. At $K=100$, 329/360 responses contained one integer (3 correct, 326 wrong), 31 were noncompliant, and 47 were truncated. (c) Correct core responses out of 1,080 (18 models $\times$ 60 items). The first two bars apply one-integer and last-integer scoring to the same core responses (16-token limit). Large budget and written steps use 512-token limits and last-integer scoring.}
\label{fig:main}
\end{figure*}

Wording sensitivity uses 30 other items, ten at each of $K=3,5,7$, rendered in five wordings (original, formal, casual, minimal, and verbose), all requesting a numerical answer.
Three protocols reuse the core items.
The instruction variant changes user and system instructions, retaining the core budget and scoring. Its system message reads \emph{You are a precise arithmetic assistant. You track numerical state changes and report final values. Always respond with only the requested number, no explanation.}
Large budget retains the core prompt with a 512-token limit.
Written steps uses the same limit but replaces the closing instruction with \emph{Think step by step, then give your final answer as a single number on the last line.}
Their last-integer comparison tests the combined instruction to write steps and place the answer on the last line (Section~\ref{sec:protocol-results}).

Intermediate-state and multi-target items extend the final-state query (Section~\ref{sec:ext-results}, Appendix~\ref{app:ext}).
\subsection{Response scoring}
\label{sec:scoring}
Core, depth, arithmetic, cancellation, wording, instruction-variant, and intermediate-state responses require exactly one extracted integer matching the executable answer (one-integer rule, Figure~\ref{fig:overview}). Zero or multiple integers receive zero credit.
Written steps uses the last integer (last-integer rule). Large budget uses both rules.
Multi-target responses must give exactly $M$ integers matching the reference values in the requested order (Appendix~\ref{app:ext}).
The parser removes dollar signs and supported thousands separators before extraction but allows surrounding prose, so compliance differs from a bare-number answer.
We report correct, one-integer-wrong, and noncompliant responses.
Truncation, the service's length finish reason, can overlap any outcome.
All core items remain in the denominator.

Assignment and set and count items use two parsers, reported in parallel with no primary parser.
Parser 2 requires one distinct correct vocabulary member for assignment, and one correct integer or exact set equality for sets and counts.
Parser 1 permits more substring, numerical-position, and empty-set matches.
Appendix~\ref{app:scoring} specifies both parsers and their seven differing labels.

\subsection{Models and external criterion}
\label{sec:execution}
\begin{table*}[t]
\centering
\footnotesize
\setlength{\tabcolsep}{3pt}
\begin{tabular}{@{}l>{\columncolor{black!9}}rrrrrrrrr@{}}
\toprule
Model & \shortstack{\textbf{Core}\\60} & \shortstack{Depth\\140} & \shortstack{Arithmetic\\90} & \shortstack{Cancellation\\100} & \shortstack{Assignment\\90 (P1/P2)} & \shortstack{Sets/counts\\90 (P1/P2)} & \shortstack{Wordings\\150} & \shortstack{Instructions\\60} & \shortstack{Written steps\\60} \\
\midrule
Granite-4.0-350M & \textbf{5} & 3 & 59 & 12 & 53/53 & 11/11 & 12 & 5 & 20 \\
Granite-3.1-8B & \textbf{15} & 7 & 76 & 69 & 84/84 & 41/41 & 56 & 17 & 54 \\
Llama-3.2-1B & \textbf{0} & 2 & 58 & 56 & 50/50 & 24/24 & 17 & 10 & 19 \\
Llama-3.2-3B & \textbf{11} & 8 & 87 & 47 & 90/90 & 42/38 & 43 & 7 & 57 \\
Llama-3.3-70B & \textbf{31} & 5 & 90 & 95 & 90/90 & 64/64 & 142 & 39 & 60 \\
Falcon3-1B & \textbf{2} & 3 & 13 & 51 & 50/49 & 6/6 & 17 & 2 & 21 \\
Falcon3-3B & \textbf{2} & 2 & 19 & 27 & 83/83 & 32/31 & 32 & 14 & 59 \\
Falcon3-7B & \textbf{17} & 8 & 88 & 87 & 90/90 & 44/44 & 54 & 15 & 60 \\
Falcon3-10B & \textbf{20} & 6 & 89 & 84 & 89/89 & 50/50 & 87 & 18 & 60 \\
Qwen3-4B-2507 & \textbf{1} & 0 & 84 & 55 & 87/87 & 49/49 & 22 & 0 & 60 \\
\specialrule{0.3pt}{1pt}{1pt} GPT-5.2 & \textbf{56} & 21 & 90 & 99 & 90/90 & 74/74 & 148 & 57 & 60 \\
GPT-4.1 & \textbf{49} & 20 & 90 & 99 & 90/90 & 71/71 & 146 & 55 & 60 \\
GPT-4.1-mini & \textbf{36} & 6 & 90 & 100 & 89/89 & 73/73 & 128 & 39 & 60 \\
Claude Haiku 4.5 & \textbf{39} & 9 & 90 & 100 & 90/90 & 67/67 & 146 & 41 & 60 \\
Claude Sonnet 4.5 & \textbf{0} & 0 & 90 & 3 & 77/77 & 65/65 & 71 & 11 & 60 \\
Claude Opus 4.5 & \textbf{14} & 0 & 90 & 39 & 90/90 & 74/74 & 103 & 52 & 60 \\
Command A & \textbf{31} & 9 & 90 & 98 & 90/90 & 61/61 & 125 & 36 & 60 \\
Command R7B & \textbf{13} & 2 & 83 & 41 & 76/76 & 29/28 & 49 & 12 & 60 \\
\bottomrule
\end{tabular}
\caption{Correct counts ($\uparrow$) across the nine main-bank blocks (total items in column headers), with all 840 requests complete for every model. Assignment and Sets/counts give two correct counts as Parser 1 / Parser 2 (P1 / P2), each out of 90. Neither is primary. Wordings uses five phrasings of the same 30 items. Instructions and Written steps reuse the 60 core items. The latter uses last-integer scoring. Scores are not pooled across blocks.}
\label{tab:full-scores}
\end{table*}

All 18 models answered all 840 main-bank requests, with no missing responses.
Ten ran locally and eight through APIs (Appendix~\ref{app:execution}, scores in Table~\ref{tab:full-scores}).
GPT-5.2 used reasoning effort \texttt{none}, and the other models used temperature zero without a reasoning mode.
No saved response reports reasoning tokens or traces.
Local runs of the main bank took 1.54 device-hours, and its API responses cost about US\$4.

We match models to released BFCL V4 scores~\citep{patil2025bfcl} from the public 16 December 2025 function-calling snapshot, without rerunning BFCL, reporting Overall, Multi-Turn, and Agentic in parallel.

\subsection{Statistical comparisons}
\label{sec:statistics}
For each BFCL criterion, we compute tie-preserving Kendall $\tau_b$ with core, single-step arithmetic, and cancellation scores.
Paired differences (core minus arithmetic and core minus cancellation) assess whether core associations exceed those of the simpler controls.
We report pointwise 95\% percentile intervals from 10,000 paired model-bootstrap samples, shared across all associations within a panel, including the instruction variant, large budget, and written steps.

A Pearson check repeats core, arithmetic, and cancellation comparisons using correct counts rather than ranks (Table~\ref{tab:protocol-pearson}).
We also leave out each of seven provider groups in turn to assess provider influence. The main analysis retains all 18 models.

Under both parsers, we compute associations of sets and counts with the core and three BFCL criteria, and of each domain with the core.
Intermediate-state and multi-target means and paired differences also use 10,000 model-bootstrap samples.
Seed, wording, and depth analyses report descriptive accuracies and rank correlations without significance tests.
Appendix~\ref{app:control-tables} gives implementation details and Pearson results. The supplementary material lists every estimate and per-model score (Appendix~\ref{app:execution}).

\section{Results}
\label{sec:results}

Core accuracy was lower than single-step arithmetic accuracy for every model, 31.7\% versus 84.9\% pooled (Figure~\ref{fig:main}a). All models remain in the analyses, including low-scoring and frequently truncated ones (Table~\ref{tab:full-scores}). Larger models scored higher in five of six adjacent-size core comparisons within Granite, Llama, and Falcon, and tied in one. Generation differences in Granite and Llama prevent isolating parameter count.

\subsection{Performance across sequence depth}
\label{sec:depth-results}

Accuracy fell from 44.4\% at $K=3$ to 0.8\% at $K=100$ (Figure~\ref{fig:main}b), where most answers were well formatted but wrong, so long-sequence failure also occurred within the scoring format. Core accuracy was 52.4\%, 32.2\%, 21.5\%, and 18.2\% with one, two, three, and four or more updates to the queried entity.

From $K=50$ onward, no model exceeded 2/20 correct.

Sonnet 4.5 truncated all 140 depth responses, so its depth scores reflect truncation.

Items are regenerated across depths, and $K$ counts updates to all entities, so these curves do not isolate appending updates to identical items.

\subsection{Intermediate states and multiple targets}
\label{sec:ext-results}
\begin{table*}[t]
\centering
\footnotesize
\setlength{\tabcolsep}{3pt}
\begin{minipage}[t]{0.49\textwidth}
\vspace{0pt}
\centering
\begin{tabular}{@{}lrrr@{}}
\toprule
Condition & Correct $\uparrow$ & Accuracy (\%) $\uparrow$ & 95\% interval \\
\midrule
State 1 & 192/360 & 53.3 & [36.1, 70.3] \\
State 2 & 131/360 & 36.4 & [17.8, 55.6] \\
State 3 & 108/360 & 30.0 & [13.3, 48.1] \\
\addlinespace[2pt]
Tracking, $M=1$ & 103/360 & 28.6 & [11.7, 47.2] \\
Tracking, $M=2$ & 91/360 & 25.3 & [8.6, 44.2] \\
Tracking, $M=4$ & 84/360 & 23.3 & [5.6, 42.8] \\
\addlinespace[2pt]
Readout, $M=1$ & 136/360 & 37.8 & [17.8, 58.3] \\
Readout, $M=2$ & 179/360 & 49.7 & [30.3, 68.9] \\
Readout, $M=4$ & 186/360 & 51.7 & [33.1, 70.0] \\
\bottomrule
\end{tabular}
\end{minipage}\hfill
\begin{minipage}[t]{0.49\textwidth}
\vspace{0pt}
\centering
\begin{tabular}{@{}lrr@{}}
\toprule
Contrast & Difference (pp) & 95\% interval \\
\midrule
State $2-1$ & $-$16.9 & [$-$27.2, $-$7.2] \\
State $3-2$ & $-$6.4 & [$-$16.7, 2.5] \\
State $3-1$ & $-$23.3 & [$-$35.8, $-$10.6] \\
Slot-position & 4.5 & [$-$0.7, 10.2] \\
\addlinespace[2pt]
R$-$T, $M=1$ & 9.2 & [1.1, 20.0] \\
R$-$T, $M=2$ & 24.4 & [11.7, 39.2] \\
R$-$T, $M=4$ & 28.3 & [15.0, 43.1] \\
\addlinespace[2pt]
T, $M=4$ minus $1$ & $-$5.3 & [$-$9.2, $-$1.9] \\
R, $M=4$ minus $1$ & 13.9 & [5.0, 24.4] \\
\bottomrule
\end{tabular}
\end{minipage}
\caption{Intermediate-state and multi-target results (18 models). Pooled correct counts and mean model accuracy (left), paired differences in percentage points (right). States 1--3 index the queried entity's updates. $M$ counts queried entities, and T/R denote tracking/readout. Positive differences favor the first term. Slot-position compares queries where $\lceil s/3\rceil$ correctly counts target updates with the rest (first minus second), where $s$ is the queried slot. Intervals are pointwise 95\% percentile model-bootstrap intervals (10,000 resamples).}
\label{tab:ext-summary}
\end{table*}

\begin{figure}[!h]
\centering
\includegraphics[width=\columnwidth]{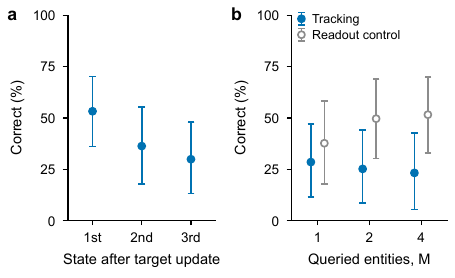}
\caption{Intermediate-state queries and multi-target items. (a) Accuracy for states after the queried entity's first, second, or third update. Each query specifies the slot. (b) Accuracy for $M=1$, 2, and 4 queried entities. Readout controls redirect target updates to other entities, leaving the queried values unchanged. The categorical axis spaces 1, 2, and 4 equally. Points are 18-model means, and error bars are pointwise 95\% percentile model-bootstrap intervals (10,000 resamples). Table~\ref{tab:ext-summary} gives values and paired differences with intervals.}
\label{fig:ext}
\end{figure}

Intermediate-state queries test end-state shortcuts by asking for earlier values (Figure~\ref{fig:ext}a). Twenty trajectories, each with five entities, contain nine slots, with three same-sign one-point updates to the queried entity, four updates to others, and two inactive slots. Each trajectory supplies one query after each target update, giving 60 items. Later states were not easier to report. Mean third-update accuracy was 23.3 points [10.6, 35.8] below first-update accuracy (Table~\ref{tab:ext-summary}).

Twenty trajectories yield 120 multi-target and readout items querying $M=1$, 2, or 4 entities (Figure~\ref{fig:ext}b). Each target receives two noncancelling updates, plus four operations on unqueried entities, giving 6, 8, or 12 operations. Readout redirects target updates to other entities while preserving initial states, slot order, and operation signs and magnitudes. Queried values therefore equal their initial values. Operation counts and singular versus plural wording vary across $M$, so cross-$M$ comparisons do not isolate capacity. Tracking and readout share instructions at each $M$ (Appendix~\ref{app:ext}). The $M=1$ instruction was changed from plural to singular after three models had run and before any others. Those three were rerun, and their first runs are retained but not analyzed. Llama-3.3-70B used a hosted endpoint here, matching its local run on the 60 core items (31 correct each).

Readout exceeded tracking by 9.2 to 28.3 percentage points across $M=1$, 2, and 4, with each difference interval above zero (Table~\ref{tab:ext-summary}). Readout itself scored only 38 to 52 percent, with most errors from truncation. At $M=1$, the nine locally run models gave 95 one-integer readout answers, and 86 were wrong although no operation touched the queried entity.

Five models truncated most of their 180 intermediate-state and multi-target responses. Three other models were near ceiling on multi-target items (Table~\ref{tab:ext-by-model}). Llama-3.3-70B scored 4/60 on tracking and 53/60 on readout, whereas GPT-5.2 scored 60/60 on both. Among format-compliant responses, tracking errors were 109/212, 143/234, and 156/240 at $M=1$, 2, and 4, and readout errors were 99/235, 73/252, and 65/251. Tracking failures thus also occurred within the scoring format.

Wrong one-integer answers included 21 reporting the second state when asked for the first. When asked for the second, 47 reported the first and 19 the final state. For 41/60 queries, slot position alone gives the update count. Accuracy on these items was 4.5 points [$-0.7$, 10.2] higher than on the other 19, with an interval including zero (Table~\ref{tab:ext-summary}).

\subsection{Direct assignment, set updates, and counting}
\label{sec:ef-results}
\begin{table}[!b]
\centering
\small
\setlength{\tabcolsep}{4pt}
\begin{tabular}{@{}lrr@{}}
\toprule
Domain & Parser 1 & Parser 2 \\
\midrule
Color assignment & 502 (93.0\%) & 502 (93.0\%) \\
Location assignment & 481 (89.1\%) & 480 (88.9\%) \\
Status assignment & 475 (88.0\%) & 475 (88.0\%) \\
\midrule Meeting counts & 184 (34.1\%) & 184 (34.1\%) \\
Permissions & 344 (63.7\%) & 338 (62.6\%) \\
Inventory & 349 (64.6\%) & 349 (64.6\%) \\
\bottomrule
\end{tabular}
\caption{Correct counts and accuracy ($\uparrow$) by task domain. Each row pools 540 responses to 30 items shared by all 18 models. Both rules score the same responses, with neither chosen as the main rule.}
\label{tab:ef-domain-summary}
\end{table}

\begin{table*}[t]
\centering
\footnotesize
\setlength{\tabcolsep}{3pt}
\begin{tabular}{lrrrrrr>{\raggedleft\arraybackslash}p{20pt}>{\raggedleft\arraybackslash}p{20pt}>{\raggedleft\arraybackslash}p{20pt}>{\raggedleft\arraybackslash}p{20pt}}
\toprule
\multicolumn{1}{l}{} & \multicolumn{3}{c}{Core} & \multicolumn{3}{c}{Large budget} & \multicolumn{2}{c}{Instructions} & \multicolumn{2}{c}{Written steps} \\
\multicolumn{1}{l}{} & \multicolumn{3}{c}{16 tokens} & \multicolumn{3}{c}{512 tokens} & \multicolumn{2}{c}{16 tokens} & \multicolumn{2}{c}{512 tokens} \\
\cmidrule(lr){2-4}\cmidrule(lr){5-7}\cmidrule(lr){8-9}\cmidrule(lr){10-11}
Model & One & Last & Trunc. & One & Last & Trunc. & One & Last & One & Last \\
\midrule
Granite-4.0-350M & 5 & 5 & 0 & 5 & 5 & 0 & 5 & 5 & 2 & 20 \\
Granite-3.1-8B & 15 & 15 & 2 & 15 & 16 & 0 & 17 & 17 & 0 & 54 \\
Llama-3.2-1B & 0 & 2 & 15 & 1 & 5 & 0 & 10 & 10 & 0 & 19 \\
Llama-3.2-3B & 11 & 11 & 0 & 11 & 11 & 0 & 7 & 7 & 0 & 57 \\
Llama-3.3-70B & 31 & 33 & 1 & 31 & 35 & 0 & 39 & 39 & 0 & 60 \\
Falcon3-1B & 2 & 2 & 0 & 2 & 2 & 0 & 2 & 2 & 0 & 21 \\
Falcon3-3B & 2 & 2 & 54 & 2 & 55 & 0 & 14 & 14 & 0 & 59 \\
Falcon3-7B & 17 & 17 & 0 & 17 & 17 & 0 & 15 & 16 & 0 & 60 \\
Falcon3-10B & 20 & 20 & 0 & 20 & 20 & 0 & 18 & 18 & 0 & 60 \\
Qwen3-4B-2507 & 1 & 10 & 40 & 1 & 42 & 0 & 0 & 18 & 0 & 60 \\
GPT-5.2 & 56 & 56 & 0 & 56 & 56 & 0 & 57 & 57 & 0 & 60 \\
GPT-4.1 & 49 & 49 & 0 & 51 & 51 & 0 & 55 & 55 & 0 & 60 \\
GPT-4.1-mini & 36 & 36 & 0 & 35 & 35 & 0 & 39 & 39 & 0 & 60 \\
Claude Haiku 4.5 & 39 & 39 & 0 & 39 & 39 & 0 & 41 & 41 & 0 & 60 \\
Claude Sonnet 4.5 & 0 & 0 & 60 & 0 & 60 & 0 & 11 & 11 & 0 & 60 \\
Claude Opus 4.5 & 14 & 14 & 45 & 15 & 60 & 0 & 52 & 52 & 0 & 60 \\
Command A & 31 & 31 & 0 & 30 & 30 & 0 & 36 & 36 & 0 & 60 \\
Command R7B & 13 & 13 & 0 & 11 & 11 & 0 & 12 & 12 & 0 & 60 \\
\midrule \rowcolor{black!9} \textbf{All 18 models (of 1,080)} & \textbf{342} & \textbf{355} & \textbf{217} & \textbf{342} & \textbf{550} & \textbf{0} & \textbf{430} & \textbf{449} & \textbf{2} & \textbf{950} \\
\bottomrule
\end{tabular}
\caption{Response counts on the same 60 core items per model under four protocols. One and Last ($\uparrow$) give correct counts under the one-integer and last-integer rules. Trunc. ($\downarrow$) counts responses at the output limit. Token counts in the headers are output limits. Large budget retains the core prompt. The instruction variant changes user and system instructions. Written steps requests steps followed by the final answer on the last line.}
\label{tab:budget}
\end{table*}

Under Parser 1, assignment accuracy was 90.0\%, versus 54.1\% for sets and counts. Sets and counts ranged from 34.1\% for meeting counts to 64.6\% for inventory (Table~\ref{tab:ef-domain-summary}).
The set and count items' association with the core was 0.536 [0.211, 0.833] under Parser 1, and their BFCL associations were 0.669 to 0.708 under either parser (Table~\ref{tab:ef}, Appendix~\ref{app:ef}). The permission and inventory domains had positive associations with the core under both parsers, whereas the counting-domain interval included zero. Only seven of the 3,240 assignment, set, and count responses changed correctness between parsers.
The assignment items differ from the core in answer type, while the set and count items also differ in entity counts and some output budgets and include counting. These blocks therefore do not isolate cumulative updating from arithmetic.

\subsection{Sensitivity to the response protocol}
\label{sec:protocol-results}

On 60 shared core items across 18 models, the larger budget added 195 correct responses under last-integer scoring, and the combined instruction added 400 more, reaching 950/1,080 (88.0\%, Figure~\ref{fig:main}c, Table~\ref{tab:budget}). The instruction change thus accounted for most of the gain. Under large budget, truncation disappeared but one-integer scoring still yielded 342 correct responses. Responses containing several integers received no credit.

The user/system instruction variant increased pooled accuracy at the unchanged budget and scoring rule, with eleven models improving, two tying, and five declining (Table~\ref{tab:budget}). Because both instructions changed, the gain cannot be assigned to either alone. Written steps improved all 18 models. Bare-number answers numbered 813, 986, and 0 of 1,080 under the core, instruction, and written-steps protocols. Truncations numbered 217, 80, and 4. Relative to the core, the instruction variant had 123 wrong-to-correct gains and 35 losses, and written steps had 615 gains and 7 losses (Appendix~\ref{app:stability}).

Claude Sonnet 4.5 and Opus 4.5 scored 0/60 and 14/60 under the core protocol, truncating most responses. Both scored 60/60 under both large-budget and written-steps protocols with last-integer scoring (Table~\ref{tab:budget}). Low core accuracy alone therefore does not imply inability to solve the underlying state-tracking problems.

\subsection{Sensitivity to generated items and wordings}
\label{sec:stability-results}
\begin{table}[t]
\centering
\small
\setlength{\tabcolsep}{4pt}
\begin{tabular}{lrrr}
\toprule
Probe / contrast & Overall & Multi-Turn & Agentic \\
\midrule
\rowcolor{black!9} \textbf{Core} & \shortstack[r]{\textbf{0.449}\\[-1pt]{\scriptsize [0.042, 0.836]}} & \shortstack[r]{\textbf{0.417}\\[-1pt]{\scriptsize [0.015, 0.792]}} & \shortstack[r]{\textbf{0.449}\\[-1pt]{\scriptsize [0.066, 0.801]}} \\
Arithmetic & \shortstack[r]{0.716\\[-1pt]{\scriptsize [0.532, 0.856]}} & \shortstack[r]{0.740\\[-1pt]{\scriptsize [0.566, 0.872]}} & \shortstack[r]{0.687\\[-1pt]{\scriptsize [0.482, 0.845]}} \\
Cancellation & \shortstack[r]{0.270\\[-1pt]{\scriptsize [-0.135, 0.683]}} & \shortstack[r]{0.290\\[-1pt]{\scriptsize [-0.119, 0.700]}} & \shortstack[r]{0.296\\[-1pt]{\scriptsize [-0.133, 0.718]}} \\
\specialrule{0.3pt}{1pt}{1pt} Core $-$ Arith. & \shortstack[r]{-0.267\\[-1pt]{\scriptsize [-0.661, 0.094]}} & \shortstack[r]{-0.323\\[-1pt]{\scriptsize [-0.696, 0.024]}} & \shortstack[r]{-0.238\\[-1pt]{\scriptsize [-0.631, 0.112]}} \\
Core $-$ Cancel. & \shortstack[r]{0.179\\[-1pt]{\scriptsize [-0.092, 0.502]}} & \shortstack[r]{0.127\\[-1pt]{\scriptsize [-0.142, 0.445]}} & \shortstack[r]{0.153\\[-1pt]{\scriptsize [-0.137, 0.474]}} \\
\bottomrule
\end{tabular}
\caption{BFCL associations and paired differences for 18 models. The first three rows give Kendall's $\tau_b$, and the last two give paired $\tau_b$ differences. Brackets give pointwise 95\% intervals (Section~\ref{sec:statistics}). Arith. and Cancel. abbreviate the arithmetic and cancellation items. All 10,000 model-bootstrap estimates were valid.}
\label{tab:associations}
\end{table}

Across four core-item generation seeds, individual scores varied but rankings stayed fairly consistent, with the six pairwise correlations averaging 0.821. GPT-4.1-mini's accuracy spanned 40 percentage points. Sonnet 4.5 and Llama-3.2-1B showed zero variation because both scored zero on every seed. Seeds generate different items, not repeated decodings of one request. The correlations are descriptive and computed over the same 18 models, so they are not independent replications.

Five numeric-only wordings of the same 30 non-core items, with a 16-token limit and one-integer rule, changed scores substantially. Their ten pairwise ranking correlations averaged 0.674. Sonnet 4.5 scored 0/30 under the original wording, with all responses truncated, and 27/30 under casual wording, with 3 truncations.

\subsection{Association with BFCL}
\label{sec:association}

Across all 18 models, core accuracy was positively associated with all three BFCL criteria (Kendall's $\tau_b$ from 0.417 to 0.449), with wide pointwise 95\% intervals, all above zero (Table~\ref{tab:associations}).

All paired Kendall differences, core minus arithmetic and core minus cancellation, had intervals including zero, so the association does not single out state tracking (Table~\ref{tab:associations}). Without the four models that truncated most core responses, core minus arithmetic became positive at 0.096, 0.005, and 0.098, with intervals still including zero.
Leaving out one provider group at a time moved these estimates but kept the core's association with BFCL Overall positive, with point estimates from 0.325 to 0.702. Removing Anthropic alone moved the core-minus-arithmetic difference for Overall from $-0.267$ to 0.011. With Pearson correlations of correct counts, core minus cancellation was 0.269 to 0.336 and all three intervals were above zero. The core minus arithmetic intervals still included zero (Table~\ref{tab:protocol-pearson}). Across the 15 BFCL subcategory scores, no core-minus-arithmetic or core-minus-cancellation interval lay above zero, and the core-minus-arithmetic intervals for the non-live, live, and missing-function categories lay below zero.

\begin{table}[t]
\centering
\small
\setlength{\tabcolsep}{4pt}
\begin{tabular}{lrrr}
\toprule
Protocol & Overall & Multi-Turn & Agentic \\
\midrule
\rowcolor{black!9} \textbf{Core} & \shortstack[r]{\textbf{0.449}\\[-1pt]{\scriptsize [0.042, 0.836]}} & \shortstack[r]{\textbf{0.417}\\[-1pt]{\scriptsize [0.015, 0.792]}} & \shortstack[r]{\textbf{0.449}\\[-1pt]{\scriptsize [0.066, 0.801]}} \\
\shortstack[l]{Large budget\\one integer} & \shortstack[r]{0.431\\[-1pt]{\scriptsize [0.021, 0.821]}} & \shortstack[r]{0.399\\[-1pt]{\scriptsize [0.000, 0.778]}} & \shortstack[r]{0.431\\[-1pt]{\scriptsize [0.043, 0.787]}} \\
\shortstack[l]{Large budget\\last integer} & \shortstack[r]{0.656\\[-1pt]{\scriptsize [0.300, 0.903]}} & \shortstack[r]{0.638\\[-1pt]{\scriptsize [0.292, 0.892]}} & \shortstack[r]{0.629\\[-1pt]{\scriptsize [0.266, 0.882]}} \\
Instructions & \shortstack[r]{0.511\\[-1pt]{\scriptsize [0.182, 0.804]}} & \shortstack[r]{0.493\\[-1pt]{\scriptsize [0.180, 0.785]}} & \shortstack[r]{0.538\\[-1pt]{\scriptsize [0.242, 0.804]}} \\
Written steps & \shortstack[r]{0.633\\[-1pt]{\scriptsize [0.427, 0.784]}} & \shortstack[r]{0.626\\[-1pt]{\scriptsize [0.427, 0.765]}} & \shortstack[r]{0.719\\[-1pt]{\scriptsize [0.480, 0.868]}} \\
\bottomrule
\end{tabular}
\caption{BFCL associations ($n=18$). Kendall $\tau_b$ [pointwise 95\% model-bootstrap interval] (Section~\ref{sec:statistics}). Core and Instructions use one-integer scoring. Written steps uses last-integer scoring. Per criterion, all 10,000 resamples were valid except for Written steps (9,994).}
\label{tab:protocol-associations}
\end{table}

Protocol-specific BFCL associations are shown in Table~\ref{tab:protocol-associations}. Instructions and Written steps had higher point estimates than Core, but their paired difference intervals included zero (Table~\ref{tab:protocol-differences}).

\section{Discussion}
\label{sec:discussion}

Core accuracy lagged arithmetic, and multi-target tracking lagged readout (Table~\ref{tab:full-scores}, Figure~\ref{fig:ext}).
With the sequence visible, final answers cannot distinguish state maintenance from recomputation or establish equivalence to human span tasks.

Changing instructions altered scores on identical items, with the combined instruction accounting for most of the protocol's gain (Figure~\ref{fig:main}c).
Low core accuracy is therefore not a condition-invariant capacity estimate. Test conditions must accompany scores.

Positive BFCL associations connect core performance to an external criterion, but uncertain differences from simpler tasks limit their interpretation (Table~\ref{tab:associations}).
The probe is better used to locate answer degradation under reported conditions than as a substitute for tool-use evaluation.

\section{Conclusion}
\label{sec:conclusion}
We introduced WMIP to evaluate cumulative state tracking through programmable tasks, executable reference answers, and protocol comparisons on identical items.
Across 18 LLMs, strong single-step arithmetic performance did not ensure reliable cumulative tracking, and accuracy declined as update sequences grew.
Positive associations with public BFCL scores further connected core performance to an external measure of tool-use performance.
Changing the response protocol substantially altered accuracy on the same problems, with the combined instruction contributing more improvement than the larger output budget.
Together, these findings show that measured tracking performance reflects both cumulative task demands and response conditions.
By making these factors configurable and comparisons reproducible, WMIP provides a framework for investigating where state-tracking accuracy degrades and how output conditions affect performance.

\section*{Limitations}
\label{sec:limitations}

The visible sequence and final answers cannot distinguish state maintenance from recomputation or establish equivalence to human working-memory tasks.
Scores depend on instructions, formatting, and output limits. One-integer responses do not rule out written steps or hidden computation.
Protocol changes are bundled, while depth and control comparisons also vary task content.
The 18-model convenience panel limits generalization. Model-bootstrap intervals omit item-sampling uncertainty and provider dependence and are not adjusted for multiplicity.
BFCL is the sole external criterion, based on a historical snapshot with partly documented settings. We report associations without matched reruns, held-out predictive evaluation, or comparisons with published probes under their official protocols.

\bibliography{main}
\appendix
\setcounter{table}{0}
\renewcommand{\thetable}{A\arabic{table}}
\renewcommand{\dbltopfraction}{0.9}\renewcommand{\topfraction}{0.95}\renewcommand{\textfraction}{0.05}\renewcommand{\floatpagefraction}{0.75}\renewcommand{\dblfloatpagefraction}{0.75}\setcounter{dbltopnumber}{1}\setcounter{topnumber}{4}\setcounter{totalnumber}{6}
\section{Task Generation and Prompts}
\label{app:prompts}
Core initial values are integers from 5 to 20.
Gains draw from 1 to 10. Losses draw from 1 to 5 subject to leaving at least one point. Transfers draw from 1 to 3 subject to the giver retaining at least one point.
The implementation substitutes a one-point gain or a no-transfer operation when the corresponding sampled update is unavailable.
For core depth $K$, generation seed $s$, and probe index $i$, the random stream uses $s+1000K+i$.
Each depth has four seed cells, each retaining five nondegenerate items. Replacement indices advance within the same cell.
The item records in the supplementary material retain the original cell slot and replacement index.

Section~\ref{sec:generation} gives one complete frozen core prompt.

The instruction variant begins with \emph{Track the following point updates carefully.}, labels the final query \emph{Question:}, and ends with \emph{Answer with ONLY the number.}
Section~\ref{sec:bank} gives its system message and the written-steps instruction.
Both protocols link to the core items by source item identity, not by output order.

The wording variants use a separate seed bank of 30 instances, ten each at $K=3,5,7$.
These items use gains and losses only and were not redrawn. In 7 of the 30, no update involves the queried entity, so the answer equals its initial value.
All five templates share each item's initial values, updates, query, and reference answer.
The minimal template appends \emph{Respond with ONLY the final number.} The other four retain their numerical-answer instructions.

\section{Scoring Specifications}
\label{app:scoring}
The numerical parser removes dollar signs and supported three-digit thousands separators, then extracts integer substrings matching the regular expression \texttt{-?\textbackslash d+}. The core rule requires exactly one match equal to gold, while the written-steps protocol uses the final match.
The bare-number diagnostic additionally removes supported markdown markers and a final period and requires the remaining string to be a single integer.
Scoring uses the complete recorded response, including one ending at the length limit.

\paragraph{Parsers for the assignment items.}
Parser 1 removes complete \texttt{<think>} segments and lowercases the response, accepting gold as a substring or as the normalized first word.
Parser 2 removes asterisks/backticks, surrounding space, and a final period, then requires exactly one distinct vocabulary member, found at word boundaries, equal to gold.
Repetitions of that member and text outside the vocabulary can still pass.

\paragraph{Parsers for the set and count items.}
For counts, Parser 1 removes complete \texttt{<think>} segments and accepts gold if it equals the first or last unsigned digit string, while Parser 2 uses the one-integer rule.
For sets, Parser 1 lowercases, splits on commas, and deletes standalone \emph{no permissions} or \emph{nothing}.
Empty-set gold also passes if the response contains \emph{no } or \emph{nothing}.
Parser 2 normalizes markup, case, surrounding space, and final periods, splits on commas or the independent word \emph{and}, and requires exact set equality.
Only standalone \emph{no permissions}, \emph{nothing}, or \emph{none} denotes an empty set. An empty response fails.
Parser 2 does not delete \texttt{<think>} segments.

\begingroup
\widowpenalty=10000
Table~\ref{tab:scoring-differences} lists all seven disagreements. Five are incorrect or contradictory permission responses, one is a spelling or word-boundary difference, and one is an empty-set response with redundant negative phrases.\par
\endgroup
\begin{table*}[t]
\centering
\footnotesize
\setlength{\tabcolsep}{3pt}
\begin{tabular}{@{}>{\raggedright\arraybackslash}p{3cm}>{\raggedright\arraybackslash}p{2.6cm}>{\raggedright\arraybackslash}p{5.2cm}>{\raggedright\arraybackslash}p{\dimexpr\textwidth-10.8cm-6\tabcolsep\relax}@{}}
\toprule
\textbf{Model and item} & \textbf{Reference answer} & \textbf{Complete response} & \textbf{Disagreement reason} \\
\midrule
Falcon3 1B\newline E-location-k7-01 & "attic" & atattic & P1 accepts a substring without word boundaries. \\
Falcon3 3B\newline F-permissions-k3-06 & [] & read, no permissions & P1 accepts a contradictory permission answer. \\
Llama 3.2 3B\newline F-permissions-k3-06 & [] & admin, no write access & P1 treats a partial denial as an empty set. \\
Llama 3.2 3B\newline F-permissions-k5-05 & [] & no execute access, no permissions, no write access & P2 rejects a valid empty-set description. \\
Llama 3.2 3B\newline F-permissions-k5-09 & ["admin", "execute"] & admin, execute, no permissions & P1 ignores the contradictory ``no permissions'' phrase. \\
Llama 3.2 3B\newline F-permissions-k7-04 & ["execute"] & execute, no permissions & P1 ignores the contradictory ``no permissions'' phrase. \\
Command R7B\newline F-permissions-k7-02 & ["admin"] & admin, no permissions & P1 ignores the contradictory ``no permissions'' phrase. \\
\bottomrule
\end{tabular}
\caption{All seven scoring disagreements among 3,240 assignment, set, and count responses. Parser 1 accepts every listed response, whereas Parser 2 rejects it. [] denotes the empty set. All seven responses stopped normally.}
\label{tab:scoring-differences}
\end{table*}

\section{Execution and Model Identity}
\label{app:execution}
Local inference uses pinned weight and tokenizer revisions, vLLM 0.30.0~\citep{kwon2023vllm}, unquantized bfloat16, vendor chat templates, and temperature zero.
Small models ran on 40\,GB MIG instances of NVIDIA A100 80\,GB GPUs. The main-bank Llama-3.3-70B run used two NVIDIA RTX PRO 6000 GPUs with 96\,GB each and tensor parallelism of two.
OpenRouter API requests fix the original provider and disable fallback, and GPT-5.2 requests omit temperature.
Saved responses pass identity and provider checks.
Local repositories and frozen revisions are in the supplementary material, with technical reports for Granite 3.0~\citep{ibm2024granite3}, Llama 3~\citep{grattafiori2024llama3}, and Qwen3~\citep{yang2025qwen3}.
Vendor-inserted dates are retained.
The API models are GPT-5.2, GPT-4.1, GPT-4.1-mini, Claude Haiku 4.5, Sonnet 4.5, Opus 4.5, Command A~\citep{cohere2025commanda}, and Command R7B. Exact request identifiers and model metadata are retained in the supplementary material.
Preflight checks verify service identity and settings without selecting models by answer accuracy.
BFCL criterion inputs and available parameter counts are in the supplementary material.

Local execution used approximately 1.50 device-hours for the main bank, 0.34 for the intermediate-state and multi-target blocks, and 0.26 for the large-budget protocol, counting each MIG instance or full GPU as one device.
These totals exclude hosted runs, failed/stopped jobs, separate tests, and the superseded multi-target runs (Appendix~\ref{app:ext}).
Returned usage reports sum to approximately US\$4.032 for 6,720 main-bank responses and 40 successful preflight checks.
Across all recorded October runs, including pilots, diagnostics, failed or stopped attempts, and superseded runs, allocations totalled 1.87 A100 MIG instance-hours and 1.84 RTX PRO 6000 GPU-hours. These resource units are reported separately.
Returned API usage across these runs totalled US\$5.642. Provider GPU-hours, unitemized retries, offline CPU work, and compute preceding these runs are unavailable.
Analyses use NumPy 2.4.3 and SciPy 1.17.1.
The supplementary README documents the released artifacts, analysis outputs, execution records, detailed timing, cost accounting, and BFCL sources and mappings.

\begingroup\widowpenalty=10000
The synthetic tasks contain no personal data. Prompts and reference answers are in English, and the materials are intended for research evaluation of LLMs.
Code uses the MIT License. Data, figures, and documentation use CC BY 4.0, and BFCL score files retain Apache-2.0 with source attribution.
Granite and Qwen models use Apache 2.0, Llama models use the corresponding Llama 3.2 or 3.3 Community License Agreement, and Falcon3 models use the TII Falcon-LLM License 2.0. The supplementary README links these licenses and the hosted-service terms.\par
\endgroup

\section{Association Details and Pearson Results}
\label{app:control-tables}
\begin{table}[t]
\centering
\small
\setlength{\tabcolsep}{2pt}
\begin{tabular}{@{}lrrr@{}}
\toprule
\textbf{Block} & \textbf{Overall} & \textbf{Multi-Turn} & \textbf{Agentic} \\
\midrule
\shortstack[l]{Instructions\\$-$ Core} & \shortstack[r]{0.063\\[-1pt]{\scriptsize [-0.206, 0.345]}\\[-1pt]{\scriptsize $u=0$}} & \shortstack[r]{0.076\\[-1pt]{\scriptsize [-0.179, 0.351]}\\[-1pt]{\scriptsize $u=0$}} & \shortstack[r]{0.089\\[-1pt]{\scriptsize [-0.148, 0.353]}\\[-1pt]{\scriptsize $u=0$}} \\[2pt]
\shortstack[l]{Written steps\\$-$ Core} & \shortstack[r]{0.184\\[-1pt]{\scriptsize [-0.190, 0.559]}\\[-1pt]{\scriptsize $u=6$}} & \shortstack[r]{0.209\\[-1pt]{\scriptsize [-0.152, 0.570]}\\[-1pt]{\scriptsize $u=6$}} & \shortstack[r]{0.271\\[-1pt]{\scriptsize [-0.084, 0.632]}\\[-1pt]{\scriptsize $u=6$}} \\[2pt]
\bottomrule
\end{tabular}
\caption{Paired differences in protocol Kendall $\tau_b$ with BFCL for all 18 models. Protocol coefficients and scoring rules are in Table~\ref{tab:protocol-associations}. Tied ceiling scores for Written steps leave some resamples undefined. Minus signs denote paired differences. Pointwise 95\% percentile intervals use the same 10,000 paired model resamples as Table~\ref{tab:associations}. Differences are computed within each resample, and $u$ counts undefined resamples.}
\label{tab:protocol-differences}
\end{table}
\begin{table}[t]
\centering
\small
\setlength{\tabcolsep}{2pt}
\begin{tabular}{@{}lrrr@{}}
\toprule
\textbf{Block} & \textbf{Overall} & \textbf{Multi-Turn} & \textbf{Agentic} \\
\midrule
\rowcolor{black!9} \shortstack[l]{Core\\(16 tokens)} & \shortstack[r]{\textbf{0.477}\\[-1pt]{\scriptsize [0.057, 0.914]}\\[-1pt]{\scriptsize $u=0$}} & \shortstack[r]{\textbf{0.365}\\[-1pt]{\scriptsize [-0.028, 0.866]}\\[-1pt]{\scriptsize $u=0$}} & \shortstack[r]{\textbf{0.477}\\[-1pt]{\scriptsize [0.021, 0.922]}\\[-1pt]{\scriptsize $u=0$}} \\[2pt]
Arithmetic & \shortstack[r]{0.616\\[-1pt]{\scriptsize [0.462, 0.778]}\\[-1pt]{\scriptsize $u=0$}} & \shortstack[r]{0.535\\[-1pt]{\scriptsize [0.394, 0.715]}\\[-1pt]{\scriptsize $u=0$}} & \shortstack[r]{0.496\\[-1pt]{\scriptsize [0.356, 0.681]}\\[-1pt]{\scriptsize $u=0$}} \\[2pt]
Cancellation & \shortstack[r]{0.164\\[-1pt]{\scriptsize [-0.349, 0.789]}\\[-1pt]{\scriptsize $u=0$}} & \shortstack[r]{0.096\\[-1pt]{\scriptsize [-0.407, 0.795]}\\[-1pt]{\scriptsize $u=0$}} & \shortstack[r]{0.141\\[-1pt]{\scriptsize [-0.382, 0.774]}\\[-1pt]{\scriptsize $u=0$}} \\[2pt]
\specialrule{0.3pt}{1pt}{1pt} Core $-$ Arith. & \shortstack[r]{-0.139\\[-1pt]{\scriptsize [-0.547, 0.271]}\\[-1pt]{\scriptsize $u=0$}} & \shortstack[r]{-0.169\\[-1pt]{\scriptsize [-0.574, 0.306]}\\[-1pt]{\scriptsize $u=0$}} & \shortstack[r]{-0.019\\[-1pt]{\scriptsize [-0.490, 0.431]}\\[-1pt]{\scriptsize $u=0$}} \\[2pt]
Core $-$ Cancel. & \shortstack[r]{0.313\\[-1pt]{\scriptsize [0.073, 0.506]}\\[-1pt]{\scriptsize $u=0$}} & \shortstack[r]{0.269\\[-1pt]{\scriptsize [0.005, 0.469]}\\[-1pt]{\scriptsize $u=0$}} & \shortstack[r]{0.336\\[-1pt]{\scriptsize [0.113, 0.522]}\\[-1pt]{\scriptsize $u=0$}} \\[2pt]
\bottomrule
\end{tabular}
\caption{Pearson $r$ sensitivity of BFCL associations for correct counts, all 18 models. Minus signs denote paired differences. Arith. and Cancel. abbreviate the arithmetic and cancellation items. Pointwise 95\% percentile intervals use the same 10,000 paired model resamples as Table~\ref{tab:associations}. Differences are computed within each resample, and $u$ counts undefined resamples.}
\label{tab:protocol-pearson}
\end{table}

\begin{table}[!b]
\centering
\begin{minipage}[t]{\columnwidth}
\vspace{0pt}
\centering
\small
\setlength{\tabcolsep}{2pt}
\begin{tabular}{lrr}
\toprule
Comparison & Parser 1 & Parser 2 \\
\midrule
Sets/counts vs. core & \shortstack[r]{0.536\\[-1pt]{\scriptsize [0.211, 0.833]}} & \shortstack[r]{0.550\\[-1pt]{\scriptsize [0.228, 0.847]}} \\
Sets/counts vs. BFCL Overall & \shortstack[r]{0.695\\[-1pt]{\scriptsize [0.493, 0.860]}} & \shortstack[r]{0.708\\[-1pt]{\scriptsize [0.514, 0.874]}} \\
Sets/counts vs. BFCL Multi-Turn & \shortstack[r]{0.678\\[-1pt]{\scriptsize [0.460, 0.850]}} & \shortstack[r]{0.691\\[-1pt]{\scriptsize [0.476, 0.863]}} \\
Sets/counts vs. BFCL Agentic & \shortstack[r]{0.669\\[-1pt]{\scriptsize [0.448, 0.846]}} & \shortstack[r]{0.682\\[-1pt]{\scriptsize [0.455, 0.862]}} \\
\specialrule{0.3pt}{1pt}{1pt} Sets/counts, counting vs. core & \shortstack[r]{0.267\\[-1pt]{\scriptsize [-0.135, 0.643]}} & \shortstack[r]{0.267\\[-1pt]{\scriptsize [-0.135, 0.643]}} \\
Sets/counts, permissions vs. core & \shortstack[r]{0.544\\[-1pt]{\scriptsize [0.207, 0.830]}} & \shortstack[r]{0.530\\[-1pt]{\scriptsize [0.193, 0.816]}} \\
Sets/counts, inventory vs. core & \shortstack[r]{0.524\\[-1pt]{\scriptsize [0.248, 0.771]}} & \shortstack[r]{0.524\\[-1pt]{\scriptsize [0.248, 0.771]}} \\
\bottomrule
\end{tabular}
\caption{All 14 association estimates for the set and count items under Parser 1 and Parser 2, reported in parallel without selecting a primary parser. All correlations use 18 models and show Kendall $\tau_b$ with pointwise 95\% model-bootstrap intervals (10,000 defined resamples, seed 42). The core items always use their frozen one-integer rule. The set and count items include 60 set-update and 30 meeting-count tasks per model.}
\label{tab:ef}
\end{minipage}

\end{table}
\begin{table*}[t]
\centering
\footnotesize
\setlength{\tabcolsep}{3pt}
\begin{tabular}{l>{\raggedleft\arraybackslash}p{22pt}>{\raggedleft\arraybackslash}p{22pt}>{\raggedleft\arraybackslash}p{22pt}>{\raggedleft\arraybackslash}p{22pt}>{\raggedleft\arraybackslash}p{22pt}>{\raggedleft\arraybackslash}p{22pt}>{\raggedleft\arraybackslash}p{22pt}>{\raggedleft\arraybackslash}p{22pt}>{\raggedleft\arraybackslash}p{22pt}rr}
\toprule
 & \multicolumn{3}{c}{Intermediate states} & \multicolumn{3}{c}{Multi-target tracking} & \multicolumn{3}{c}{Multi-target readout} & & \\
 & \multicolumn{3}{c}{after update} & \multicolumn{3}{c}{$M$} & \multicolumn{3}{c}{$M$} & & \\
\cmidrule(lr){2-4}\cmidrule(lr){5-7}\cmidrule(lr){8-10}
Model & 1 & 2 & 3 & 1 & 2 & 4 & 1 & 2 & 4 & Format & Truncated \\
\midrule
Granite-4.0-350M & 3 & 0 & 0 & 1 & 0 & 0 & 0 & 4 & 0 & 151 & 0 \\
Granite-3.1-8B & 11 & 1 & 0 & 0 & 0 & 0 & 0 & 14 & 9 & 129 & 61 \\
Llama-3.2-1B & 6 & 0 & 0 & 0 & 0 & 0 & 6 & 1 & 9 & 101 & 37 \\
Llama-3.2-3B & 13 & 0 & 0 & 1 & 0 & 0 & 1 & 17 & 8 & 179 & 0 \\
Llama-3.3-70B & 19 & 20 & 7 & 4 & 0 & 0 & 20 & 14 & 19 & 122 & 46 \\
Falcon3-1B & 0 & 0 & 0 & 0 & 0 & 0 & 0 & 18 & 9 & 103 & 80 \\
Falcon3-3B & 0 & 0 & 0 & 0 & 0 & 0 & 0 & 0 & 0 & 30 & 180 \\
Falcon3-7B & 13 & 5 & 3 & 0 & 0 & 0 & 0 & 2 & 1 & 83 & 102 \\
Falcon3-10B & 18 & 16 & 8 & 4 & 3 & 0 & 2 & 7 & 17 & 180 & 0 \\
Qwen3-4B-2507 & 0 & 0 & 0 & 0 & 0 & 0 & 0 & 0 & 2 & 46 & 180 \\
GPT-5.2 & 20 & 20 & 19 & 20 & 20 & 20 & 20 & 20 & 20 & 180 & 0 \\
GPT-4.1 & 20 & 14 & 20 & 20 & 20 & 20 & 20 & 20 & 20 & 180 & 0 \\
GPT-4.1-mini & 20 & 19 & 13 & 16 & 17 & 15 & 20 & 20 & 20 & 180 & 0 \\
Claude Haiku 4.5 & 12 & 16 & 19 & 20 & 17 & 20 & 20 & 20 & 20 & 178 & 2 \\
Claude Sonnet 4.5 & 0 & 0 & 0 & 0 & 0 & 0 & 0 & 0 & 0 & 13 & 180 \\
Claude Opus 4.5 & 7 & 0 & 0 & 1 & 0 & 0 & 0 & 0 & 0 & 9 & 172 \\
Command A & 18 & 16 & 17 & 14 & 14 & 9 & 20 & 20 & 20 & 180 & 0 \\
Command R7B & 12 & 4 & 2 & 2 & 0 & 0 & 7 & 2 & 12 & 180 & 0 \\
\midrule
\rowcolor{black!9} \textbf{All 18 models (of 360)} & \textbf{192} & \textbf{131} & \textbf{108} & \textbf{103} & \textbf{91} & \textbf{84} & \textbf{136} & \textbf{179} & \textbf{186} &  &  \\
\bottomrule
\end{tabular}
\caption{Correct responses out of 20 per cell for the intermediate-state queries and the multi-target items. Intermediate-state columns give the queried update. Multi-target columns give the number of queried entities $M$ for tracking and for the matched readout control. Format counts responses with exactly the requested number of integers and Truncated counts responses that reached the output limit, both out of 180.}
\label{tab:ext-by-model}
\end{table*}

Bootstrap samples use NumPy RandomState seed 42 with common indices within each panel. For rank-based adjustments, ties receive average ranks in the original panel and each bootstrap sample. Pearson correlations use the original, unranked scores.
Undefined bootstrap statistics are counted and excluded only from the corresponding percentile calculation, never replaced by zero.
Paired differences require both components to be defined in the same draw.
The supplementary leave-one-provider-out estimates are descriptive sensitivity analyses, not replacement panels.

The Kendall comparisons establish neither an ordering nor equivalence of the core and single-step arithmetic associations (Table~\ref{tab:associations}).
Pearson estimates gave a smaller core--arithmetic gap than Kendall estimates.
Pearson differences for core minus cancellation were 0.269 to 0.336, with all three intervals above zero (Table~\ref{tab:protocol-pearson}).

\section{Assignment, Set, and Count Results}
\label{app:ef}
Neither aggregate agreement nor association magnitude is used to select a parser.
The permission and inventory domains had positive associations with the core under both parsers, whereas the counting-domain interval included zero (Table~\ref{tab:ef}).
Without correlation-difference tests or adjusted analysis, the set and count correlations do not establish stronger associations than those of the core, single-step arithmetic, or cancellation items.
Table~\ref{tab:ef-domain-summary} gives domain scores.
Overall, assignment correct counts were 1,458/1,620 (90.00\%) and 1,457/1,620 (89.94\%) for Parsers 1 and 2. Set and count totals were 877/1,620 (54.14\%) and 871/1,620 (53.77\%).
Only seven of these 3,240 responses change correctness between parsers.
The supplementary material retains all per-model domain scores under both parsers.

\section{Protocol and Stability Results}
\label{app:stability}
The large-budget protocol uses the pipeline and endpoints of the intermediate-state queries and the multi-target items, including the hosted Llama-3.3-70B endpoint (Appendix~\ref{app:ext}).
For the core, instruction, and written-steps protocols (1,080 responses each, zero missing), one-integer counts were 855, 987, and 5, respectively.
The format counts are descriptive diagnostics for written steps, not correctness gates.
Section~\ref{sec:protocol-results} reports bare-answer counts, truncations, and paired correctness changes.
The supplementary material retains complete per-model protocol diagnostics and all seed, wording, and depth scores.
The seed- and wording-pair correlations are descriptive Kendall $\tau_b$ over the same 18 models, not independent replications. No intervals or repeated-generation reliability estimates were computed.
The supplementary material retains all 16 correlations.

\section{Intermediate-state and Multi-target Blocks}
\label{app:ext}
With scoring rules and analysis specified before any of these runs, all 18 models received 180 requests each using the main-evaluation request settings.
Llama-3.3-70B used a hosted endpoint (OpenRouter, provider Novita, listed as bf16) because its two-GPU main-evaluation node was unavailable.
A system message reproduced the two date lines inserted by its default chat template, matching the local prompt token counts.
Both deployments answered 31 of the 60 core items correctly, agreeing on correctness for 56 and response text for 48.
Omitting the system message shortened prompts by 24 tokens and yielded 25 correct responses.

\paragraph{Intermediate-state queries.}
Section~\ref{sec:ext-results} describes the trajectories and paired query conditions.
Prompts end with ``Respond with ONLY the number.'' and allow 16 output tokens under the core one-integer rule.

\paragraph{Multi-target items.}
Each multi-target trajectory lists twelve entities. Section~\ref{sec:ext-results} describes the matched readout control.
Prompts request final values in a stated order, separated by commas, for $M=2$ and 4. For $M=1$, they request one final value and end with ``Respond with ONLY the number.''
The 48-token output limit applies to all $M$, and correctness requires exactly $M$ integers matching the reference values in order.
Both blocks remove names such as E0 before counting integers.

\paragraph{Change after the first responses.}
After three models answered the $M=1$ items with the other multi-target items' plural instruction, it was changed to the singular form above, before any other model ran.
Llama-3.2-3B had returned several integers on all 40 such items and Falcon3-1B on 38.
All three models were rerun on the final bank, and their first runs remain in the supplementary material but are not analyzed.
Instruction wording therefore confounds comparisons between $M=1$ and larger $M$, but not readout versus tracking at a given $M$, where both share the instruction.

\paragraph{Results.}
All 19 correct responses reaching the output limit were in readout cells and included explanatory text in addition to the requested values.

\section{Extended Related Work}
\label{app:related}
This appendix supplements Section~\ref{sec:related} with task settings, findings, and methodological comparisons.

\paragraph{Behavioral state tracking.}
For object movement, insertion, and removal, \citet{entitytracking2023} report accuracy by the number of operations affecting the queried box and test generalization to new wording and longer sequences.
On BIG-Bench Hard's object-swap task, three 2022 models achieved between 53\% and 85\% accuracy with chain-of-thought prompting, up from near chance. Scores were averaged over three, five, and seven swaps~\citep{suzgun2023challenging}.
Other tasks apply up to ten swaps, where chain-of-thought prompting also raises accuracy~\citep{rezaee2025statetracking}.
\citet{chen2024states} ask models to add a list of integers without writing steps.
They report that the strongest models directly output sums of more than ten addends and that accuracy falls as the number of addends increases.
Their task keeps one running total, whereas our core items distribute updates across three entities.
\citet{lamalfa2024code} ask for one variable's final value after short programs of additions, subtractions, and assignments.
They note that most errors occurred when a model output only the final result, not the execution trace.

\paragraph{Programmable memory tests.}
Beyond quantity updates, Minerva includes set updates and composite state tracking. Its main quantity-state evaluation uses 200 operations, with an ablation over operation count~\citep{xia2025minerva}.
GSM-Infinite reports a sigmoid decline in accuracy as operation count grows~\citep{zhou2025gsminfinite}.
Michelangelo's Latent List task pads long contexts with statements that leave the list unchanged, including blocks of operations that cancel each other~\citep{vodrahalli2024michelangelo}.
It sets complexity by the number of operations that change the answer.
Our cancellation control instead scores items built only from adjacent updates that cancel.

\paragraph{Working-memory task differences.}
Working-memory research separates holding information from using it during ongoing processing~\citep{baddeley1974working,engle1999individual}.
Operation span interleaves arithmetic judgments with words recalled later~\citep{conway2005working}.
\citet{gong2024working} give ChatGPT verbal and spatial $n$-back tasks and score match detection with $d'$. \citet{zhang2024working} apply the spatial version to models of different sizes.
\citet{hong2025exploring} place several demanding subtasks before a question and find that answer quality falls as subtasks are added.
\citet{haznitrama2026neuropsychologically} adapt neuropsychological tests and keep the move history of their spatial working-memory task in the prompt, which they argue removes the demand to maintain it.
In adults, the same arithmetic chains become much harder when operands must come from a memorized list rather than the screen~\citep{oberauer2001dissociating}.

\paragraph{Behavior and internal mechanisms.}
\citet{huang2025language} study whether models keep an unexpressed state, such as a hidden sequence of integers, whereas our core items state the initial values.
On the boxes task of \citet{entitytracking2023}, which lacks integer updates, probes and interventions on three open models indicate that they gather the relevant information at the query token rather than update the state step by step~\citep{tang2026entity}.

\paragraph{Prompt and protocol sensitivity.}
Prompt-template choices also affect comparisons of in-context learning methods~\citep{voronov2024mind}.
Heuristic scoring can inflate measured prompt sensitivity compared with judgments by a language model~\citep{hua2025flaw}.
Restricting output to formats such as JSON generally lowered accuracy on reasoning tasks but not on classification tasks~\citep{tam2024speak}.
Evaluations with more auxiliary task demands, such as producing rather than choosing an answer, can mask abilities that lower-demand evaluations reveal~\citep{hu2024auxiliary}.
Prompts that elicit written reasoning raise accuracy on multi-step problems~\citep{wei2022chain,kojima2022large}.
\citet{sprague2025cot} combine published comparisons with new experiments and find that these gains concentrate on mathematical and symbolic tasks.
Their follow-up experiments attribute the gains mainly to carrying out the computation.
Reasoning models that write long traces still collapse beyond a model-specific problem complexity on planning puzzles~\citep{shojaee2025illusion}.

\paragraph{Measurement validity.}
Construct-validity accounts relate a measure's target phenomenon to its tasks and scoring~\citep{cronbach1955construct,messick1989validity,jacobs2021measurement}.
For LLM benchmarks, \citet{bean2025measuring} additionally emphasize explicit operational definitions, analysis of unrelated task demands, and validation of automated parsing.

\paragraph{External tool-use evaluation.}
\citet{momente2025triangulating} report a rank correlation of 0.72 between accuracy on a verbal 3-back task and dialogue-game performance across nine open models and call such links speculative.
Scores on standard benchmarks lie close to a low-dimensional capability space that also predicts aggregate scores on agent benchmarks~\citep{ruan2024observational}.

\end{document}